\documentclass[11pt]{article}

\usepackage[final]{acl}

\usepackage{times}
\usepackage{latexsym}
 \usepackage{booktabs} 
\usepackage[T1]{fontenc}
\usepackage{amsmath} 
\usepackage[utf8]{inputenc}
\usepackage{microtype}
\usepackage{multirow}
\usepackage{inconsolata}
\usepackage{graphicx}
\usepackage{xcolor}
\usepackage{colortbl}
\usepackage{hyperref}

\definecolor{lightblue}{RGB}{200,220,255}
\title{SemEval-2026 Task 6: CLARITY – Unmasking Political Question Evasions}

\author{Konstantinos Thomas\textsuperscript{1}
  Giorgos Filandrianos\textsuperscript{1}
  Maria Lymperaiou\textsuperscript{1} \\
  \textbf{Chrysoula Zerva\textsuperscript{2,3,4}}\and
  \textbf{Giorgos Stamou\textsuperscript{1}} \\ 
  \textsuperscript{1}National Technical University of Athens \\
  \textsuperscript{2}Instituto de Telecomunicações \\
  \textsuperscript{3}Instituto Superior Técnico, Universidade de Lisboa\quad
  \textsuperscript{4}ELLIS Unit Lisbon \\
  \texttt{\{kthomas, geofila, marialymp\}@ails.ece.ntua.gr} \\
  \texttt{chrysoula.zerva@tecnico.ulisboa.pt, gstam@cs.ntua.gr }  \\  
  }
\begin{document}
\maketitle
\begin{abstract}
Political speakers often avoid answering questions directly while maintaining the appearance of responsiveness. Despite its importance for public discourse, such strategic evasion remains underexplored in Natural Language Processing. We introduce SemEval-2026 Task 6, CLARITY, a shared task on political question evasion consisting of two subtasks: (i) clarity-level classification into \textit{Clear Reply}, \textit{Ambivalent}, and \textit{Clear Non-Reply}, and (ii) evasion-level classification into nine fine-grained evasion strategies. The benchmark is constructed from U.S. presidential interviews and follows an expert-grounded taxonomy of response clarity and evasion. The task attracted 124 registered teams, who submitted 946 valid runs for clarity-level classification and 539 for evasion-level classification. Results show a substantial gap in difficulty between the two subtasks: the best system achieved 0.89 Macro-F1 on clarity classification, surpassing the strongest baseline by a large margin, while the top evasion-level system reached 0.68 Macro-F1, matching the best baseline. Overall, large language model prompting and hierarchical exploitation of the taxonomy emerged as the most effective strategies, with top systems consistently outperforming those that treated the two subtasks independently. CLARITY establishes political response evasion as a challenging benchmark for computational discourse analysis and highlights the difficulty of modeling strategic ambiguity in political language.
\end{abstract}

\section{Introduction}
Political theory has long conceptualized response ambiguity as a strategic means of manipulating interviews, debates, and parliamentary proceedings rather than a conversational fallacy. Evasion, referring to the deliberate action of avoiding a direct response to a question while preserving conversational legitimacy, enables politicians to effectively sidestep accountability and commitment \cite{bull-equivocation, CLAYMAN-2001}. Such practices are commonly employed in high-stakes political contexts, leaving uncertain impressions of whether the requested information was delivered while allowing for multiple interpretations of what was said.

Computational methods in modeling political speech involve identifying ideology positioning, \cite{iyyer-etal-2014-political}, strategic framing \cite{baumer-etal-2015-testing, field-etal-2018-framing}, communicative intent \cite{ferracane-etal-2021-answer}, persuasion \cite{political-persuasion} and expressed or inferred political stance \cite{multilingual-stance, Burnham-2025}. Rhetorical pathologies in political discourse have been studied through the detection of argumentative fallacies, such as invalid reasoning chains or unsupported claims \cite{jin-etal-2022-logical, mancini-etal-2024-multimodal, cantin-larumbe-chust-vendrell-2025-argumentative}. Even though fallacy detection paves the way toward capturing manipulative political practices, it presupposes the articulation of an argument that can  be evaluated for validity or soundness. Consequently, cases where the politicians strategically avoid making explicit claims altogether cannot be covered.

Nevertheless, avoiding a direct response, or disseminating information in an ambiguous way is a well-documented phenomenon in political science. Interestingly, politicians only respond explicitly to 39-46\% of the questions posed during televised interviews, in stark contrast to non-politicians, who score substantially higher in response clarity (70-89\%) \cite{bull-2003-microanalysis}. Multiple interpretations are achieved using a range of discursive strategies, including topic shifts, partial answers, and vague formulations that preserve interactional legitimacy while limiting commitment \cite{Watzlawick1967PragmaticsOH, Rasiah2010AFF}. Surprisingly, such prevalent phenomena have not received the appropriate attention by the NLP community.

To cover this gap, \citet{thomas-etal-2024-never} introduced the task of \textit{response clarity classification}, where an evasion taxonomy is proposed to cover the predominant types responsible for response ambiguity in political discourse. Due to the high importance and complexity of this unexplored direction in political discourse, we propose the shared task CLARITY, enabling the proposal and systematic comparison of multiple computational methods to advance question evasion classification.

The task comprises two tracks to evaluate two levels of granularity: i) \textbf{Clarity-level Classification}: given a question-answer pair, a system classifies the answer in three coarse-grained labels, namely \textit{Clear Reply}, \textit{Ambivalent} or \textit{Clear Non-Reply}. ii) \textbf{Evasion-level Classification}: given a question-answer pair, a system classifies the answer into one of the 9 fine-grained evasion techniques, namely \textit{Explicit}, \textit{Implicit}, \textit{Dodging}, \textit{General}, \textit{Deflection}, \textit{Partial}, \textit{Declining}, \textit{Claims Ignorance}, \textit{Clarification}. The two levels are not independent and the taxonomy is fully adopted from \citet{thomas-etal-2024-never}. The evaluation metric is Macro-F1 score, ensuring balanced performance across all classes.

A total of 124 teams registered for the shared task, of which 30 chose to submit a system description paper. During the testing and evaluation phases, participants collectively submitted 946 valid prediction sets for the clarity-level subtask and 539 for the evasion-level subtask. We interpret this level of participation, together with the diversity of methodological approaches proposed by the teams, as strong indicators of the task’s success. The substantial number of participants also enables us to identify and analyze key trends in the strategies used to address the task. 

Out of the 41 teams submitting results to the leaderboard (regardless of consequent paper submission), seven of them managed to achieve comparable or better performance than the best baseline provided for clarity-level classification, with the winning team advancing Macro-F1 score by 7pp, from 0.82 to 0.89. Similarly, on evasion-based classification seven teams out of the 33 submitted managed to achieve performance better than the provided baseline (0.57) with the rest of the teams scoring below. Leading systems primarily rely on LLMs, accompanied by advanced prompting strategies that enable capturing discriminative features for each category, as well as exploitation of the taxonomical relationships between evasion and clarity categories. 

\section{Related work}
\paragraph{Political theory}
Research in political communication has long established non-straightforward utterances as a countermeasure to accountability rather than as a deviation from cooperative norms, relying on response under-specification and strategic ambiguity \cite{equivocation}. Evasive strategies are learned and institutionalized \cite{bull-not-to-answer}, serving as rational actions against communicative conflict and public pressure \cite{bull-equivocation}. Techniques such as ambiguity, verbosity, and rhetorical padding delay commitment remarks within political exchanges \cite{bull-2003-microanalysis}. At the same time, responsiveness and evasion share a very subtle boundary shaped by the structural properties and interactional norms governing political questioning \cite{CLAYMAN-2001}. Over time, evasive techniques have been normalized in formal political discourse, underscoring their deliberate use over lack of adequate capacity \cite{bull-cant-answer}. 

\paragraph{Political QA} has explored how politicians respond to public questioning, with an emphasis on response intent, answer quality, and informational adequacy. Prior work studied the intentions underlying response formulation and subjectivity of interpretations in the context of congressional hearings \cite{ferracane-etal-2021-answer}, as well as assessing the quality or informativeness of political answers using automated evaluation methods \cite{political-qas-quality}. QA  interactions have been systematically compiled in large-scale datasets including congressional exchanges \cite{rudra2025cquericongressionalquestionsexchanges} and structured committee hearing transcripts for modeling interactional dynamics \cite{hiray-etal-2024-cocohd}. Prompt-based approaches further investigate clarity evaluation in political QA \cite{prahallad2026promptbasedclarityevaluationtopic}, motivating benchmarks that explicitly operationalize evasiveness and ambiguity.

\paragraph{Answerability and ambiguity} in responses constitute main concerns of political discourse analysis, shifting the focus from modeling question uncertainty \cite{sulem-etal-2021-know-dont, unanswerable} to gauging whether the response provides the information requested. Human judgments in such setups can be inherently many-sided, reflecting multiple perspectives that can legitimately coexist for the same input \cite{baan2022stop, plank2022problem}, while in some cases,  no candidate response adequately addresses the question at all \cite{ReyesMontesinos2025}. Contrary to everyday discourse, political settings are often associated with deliberate non-answerability and strategic ambiguity, as manifested through evasion patterns which are found to govern political interviews \cite{thomas-etal-2024-never}.

\section{Data}
\subsection{Training \& Validation Data}
\subsubsection{Data collection}
CLARITY includes presidential interviews provided by the official Whitehouse website from 4 U.S. presidents: George W. Bush, Barack Obama, Donald J. Trump, Joseph R. Biden. In total,  287 unique interviews
spanning from 2006 until 2023 were gathered.
To create QA pairs from the dialog, we employ ChatGPT (gpt3.5-turbo) to decompose multi-barreled questions into sub-question and sub-response pairs, resulting in a total of 3,756 question-response pairs, 3,448 of which were the train set and 308 the validation set. An overview of the dataset is provided in Table~\ref{tab:dataset_stats}.

\subsubsection{Data Annotation}
We employ 3 human annotators together with an expert with a background in political science who ensures the validity of the annotation process. They are provided with the original text of the political interview before QA decomposition, as well as with the occurring sub-QA pairs. To ensure the validity of the decomposed QA pairs, the annotators are tasked to both consult the original text and the decomposed samples and report any inconsistencies stemming from ChatGPT. We found this procedure to be more efficient than tasking the annotators to manually extract sub-QA pairs, especially taking into account the difficulty of the task and the time needed to decide on a final classification label.

Ultimately, the annotators perform the three following tasks: 1) Evaluate the validity of QA decomposition from ChatGPT, 2) label each decomposed QA pair according to the taxonomy, and 3) add any missing QAs and assign a label to them. As a result, each annotator evaluated $\sim$1150 samples.

\subsubsection{Annotators’ Reliability}
Annotators are first acclimatized to the task during a ``training'' period, studying already expert-annotated samples from all the categories and passing a quiz to ensure adequate comprehension of the fundamentals. Since decomposed QAs are an important aspect to be validated, we demonstrate successful versus unsuccessful cases of decomposition. 
The training proceeds with daily sessions by the expert, demonstrating the particularity of the taxonomy labels and validating comprehension via short quizzes; upon successful completion, the annotators are ready to proceed with actual, non-annotated samples. Throughout the annotation process, the expert performs weekly checks on the annotations to detect any possible drifts.
To certify that they cautiously consult the original full text along with the decomposed QAs and not the simpler QAs alone, we inject 31 additional samples containing \textit{counterfactual QAs}. These are deliberately manipulated using ChatGPT to guide classification towards a different label from the intended one. Upon generation of the counterfactual QAs, we manually revise them to ensure that successful misleading is achieved. The annotators successfully passed the counterfactual test, selecting the proper label in these cases, thus ensuring that they both consult the original text and the extracted QAs.

\subsubsection{Annotation agreement}
\begin{figure}[h!]
    \centering
    \includegraphics[width=0.63\columnwidth]{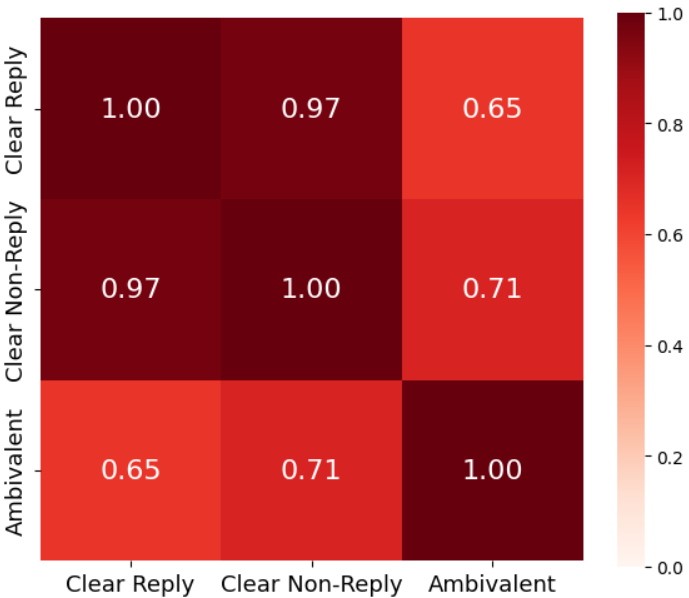}
    \caption{Annotators' agreement using Fleiss $\kappa$ for labels assigned to the clarity classification  level.}
    \label{fig:clarity-heatmap}
\end{figure}

\begin{figure}[h!]
    \centering
    \includegraphics[width=0.99\columnwidth]{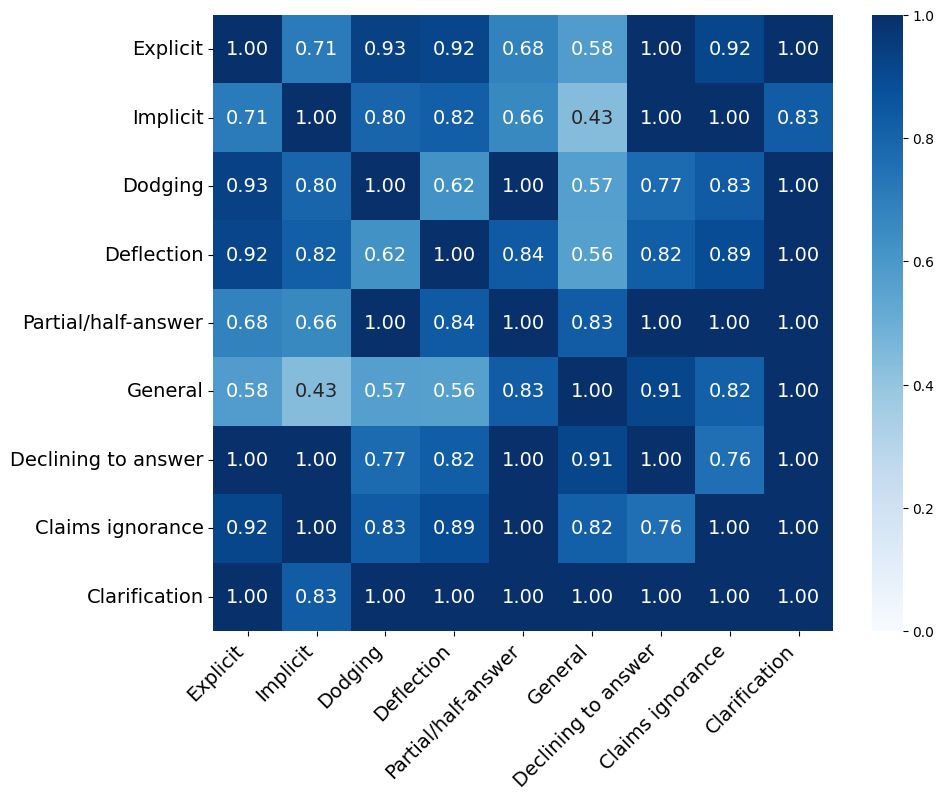}
    \caption{Annotators' agreement using Fleiss $\kappa$ for labels assigned to the evasion classification  level.}
    \label{fig:evasion-heatmap}
\end{figure}

To guarantee the best possible consistency among annotators' perception, we isolate 317 QA pairs as validation set, which receive overlapping annotations from all 3 annotators. Then, we calculate the inter-annotator agreement for both  clarity and evasion levels of the taxonomy using \textit{Fleiss} $\kappa$ \textit{score} \cite{fleissk}. Details about the mathematics of the metric are provided in App. \ref{sec:fleiss}.
Fleiss $\kappa$ values are interpreted within the following ranges:
\begin{itemize}
\item $\kappa < 0.00$: Poor agreement
\item $0.00 \le \kappa \le 0.20$: Slight agreement
\item $0.21 \le \kappa \le 0.40$: Fair agreement
\item $0.41 \le \kappa \le 0.60$: Moderate agreement
\item $0.61 \le \kappa \le 0.80$: Substantial agreement
\item $0.81 \le \kappa \le 1.00$: Almost perfect agreement
\end{itemize}
In case disagreements between annotators occur, the `gold label' is decided using majority voting. In the edge case where all three disagree, the expert selects the `gold label'. Confusion matrices regarding annotators' agreement on the clarity and evasion level are presented in Figures \ref{fig:clarity-heatmap}, \ref{fig:evasion-heatmap}, outlining the varying characteristics of categories: category pairs scoring lower $\kappa$ scores denote diverging perspectives and personal biases of the annotators, which are however increasingly valuable to the NLP community \cite{baan2022stop, plank2022problem}. 

The clarity level exhibits moderate to high agreement overall, ranging between $0.65\leq \kappa \leq 1.00$. Specifically, distinguishing between \textit{Clear Reply} and \textit{Clear Non-Reply} proves largely straightforward, yielding a near-perfect agreement ($\kappa$=0.97). In contrast, \textit{Ambivalent} replies are harder to identify, resulting in comparatively more mediocre agreement scores with the other two categories ($\kappa$=0.65 for \textit{Ambivalent} vs \textit{Clear Non-Reply}, $\kappa$=0.71 for \textit{Ambivalent} vs \textit{Clear Reply}).

The evasion level proves significantly though expectedly harder. We can identify some clear separation between categories that yield high $\kappa$ scores, such as \textit{Declining to Answer} vs \textit{Explicit}, \textit{Claims Ignorance} vs \textit{Implicit}, \textit{Partial} vs \textit{Dodging} and \textit{Clarification} vs almost any other category except \textit{Implicit}. On the other hand, label pairs of lower agreement denote the increased confusion from the annotators' side; such pairs include \textit{General} vs \textit{Implicit} ($\kappa$=0.43), \textit{General} vs \textit{Deflection} ($\kappa$=0.56), \textit{Partial} vs \textit{Implicit} ($\kappa$=0.66) among others.
\subsection{Evaluation Data}
Since the full training and validation data were already published in \citet{thomas-etal-2024-never}, we created a new test dataset to serve as evaluation and ensure unbiased assessment. To this end, we conducted an additional round of QA extraction from presidential interviews, yielding the evaluation set. This set comprises 237 QA pairs sourced from 43 interviews, annotated via the same pipeline as before but with two annotators per pair due to time and resource constraints. The evaluation dataset was opened to the participants on the day the Evaluation Phase began. The three splits of the dataset can be seen in Table~\ref{tab:dataset_stats}.

\begin{table}[t!]
\centering
\small
\begin{tabular}{l c}
\toprule
\textbf{Split} & \textbf{\# QA Pairs} \\
\midrule
Training & 3448 \\
Validation & 308 \\
Test (Evaluation Phase) & 237 \\
\bottomrule
\end{tabular}
\caption{Number of question–answer (QA) pairs in each split of the CLARITY dataset.}
\label{tab:dataset_stats}
\end{table}

\section{Task Description}
Participants are provided with textual QA pairs and are tasked to classify each in the appropriate category per subtask. They are allowed to participate in either subtask independently or both. We choose Codabench as the competition platform for both the development and evaluation phases.
\subsection{Subtasks}
\subsubsection{Subtask 1: Clarity-level Classification}
The goal of Subtask~1 is to determine the \textit{clarity level} of a response with respect to the posed question. Given a question–answer (QA) pair extracted from a presidential interview, systems must classify the response into one of three categories describing how clearly the question is addressed.

This subtask focuses on the overall communicative clarity of the response rather than the specific rhetorical strategy employed. It therefore represents a coarse-grained categorization of answer relevance and informativeness. The task is formulated as a 3-class classification problem over QA pairs.

\subsubsection{Subtask 2: Evasion-level Classification}
Subtask 2 focuses on identifying the specific \textit{evasion strategy} used in the response. Given the same QA pairs, participants must assign the response to one of fine-grained categories that capture how political speakers avoid providing a direct answer.

This subtask is formulated as a multi-class classification problem where each QA pair is assigned a single evasion category from the taxonomy of \citet{thomas-etal-2024-never}. While Subtask~1 captures whether a response clearly addresses a question, Subtask~2 characterizes the specific mechanism of evasion when a direct answer is not provided.

\subsection{Evaluation Metrics and Baselines}
Both subtasks utilize Macro-F1 score for evaluation of $k$ distinct categories. 
\begin{equation}
 \mathrm{Macro\text{-}F1}
= \frac{1}{k} \sum_{c=1}^{k} \mathrm{F1}_c   
\end{equation}
A prediction is considered correct if it matches any annotator's label.

The baselines, mentioned in \citet{thomas-etal-2024-never}, are separated in prompting and tuned approaches; the best performing variants from each approach category are summarized in Table \ref{tab:baselines}.

\begin{table}[t!]
\centering
\small
\begin{tabular}{l c c}
\toprule
\textbf{Taxonomy} & \textbf{Model} & \textbf{Macro-F1} \\
\midrule
\textit{Clarity-level (Subtask 1)} & Llama-70b & 0.82 \\
\textit{Evasion-level (Subtask 2)} & Llama-70b & 0.57 \\
\bottomrule
\end{tabular}

\caption{Baselines for clarity and evasion levels, reached by a fine-tuned Llama 70b model.}
\label{tab:baselines}
\end{table}

\section{Participating Systems and Results}
\subsection{Overview}
A total of 30 teams submitted system description forms detailing their approaches to one or both subtasks of the CLARITY shared task. Participation was higher for Subtask 1, with 29 teams submitting results for clarity-level classification, 18 of which attempted also the more challenging fine-grained evasion classification in Subtask 2. One team targeted solely Subtask 2. The participating teams represented a wide range of institutions and backgrounds, including universities and research labs across the world. Approaches spanned the full methodological spectrum, from \textbf{classical machine learning} with handcrafted features, through \textbf{fine-tuned encoder-only transformer model}s, to sophisticated \textbf{multi-stage pipelines built on frontier LLMs},  reflecting the breadth of techniques currently applied to discourse-level NLP tasks. The following sections describe and analyze the systems submitted for each subtask in turn, examining both the overall landscape of approaches and the characteristics of the most successful systems.

\subsection{Subtask 1 - Clarity-level}
\subsubsection{Overview and Participation}
Subtask 1 required participants to perform a three-way classification of question–answer (QA) pairs from presidential interviews into Clear Reply, Ambivalent, or Clear Non-Reply. A total of 29 teams submitted systems for this subtask, making it the more widely attempted of the two subtasks. The top-scoring team reached 0.89 Macro-F1, surpassing the 0.82 fine-tuned baseline. The complete leaderboard is shown in Table \ref{tab:leaderboards}.

\subsubsection{Dominant Approach: LLM-Based Prompting}
The most striking pattern in the Subtask 1 results is the \textbf{dominance of LLM-based approaches} at the top of the leaderboard. All five top-ranked systems (scores 0.82–0.89) relied primarily on LLMs with sophisticated prompting strategies rather than fine-tuned encoder-only models. This stands in notable contrast to many participants in the middle and lower tiers, who relied on smaller fine-tuned encoders such as DeBERTa and RoBERTa.

The top-ranked system \textbf{did not directly classify} Subtask 1 at all. Instead, the team solved Subtask 2 first using a three-stage dynamic prompting pipeline and then mapped the resulting fine-grained evasion labels back to the corresponding clarity categories using the task's official taxonomy hierarchy. Particularly, they followed the inverse process of \citet{thomas-etal-2024-never}, where evasion classification drove consequent clarity classification, leading to improved scores compared to independent classification on each taxonomy level. The winning reverse derivation strategy proved to be the \textbf{single most effective approach} across the entire competition, suggesting that reasoning about fine-grained evasion types is inherently more tractable than reasoning about coarse clarity labels in isolation.

The following two systems achieved score parity despite a fundamentally different design philosophy. AsymVerify designed an \textbf{asymmetric confidence-gated system} using GPT-5.2 with three verification passes: an initial classification, a downgrade pass that re-examined Clear Reply and Clear Non-Reply predictions for ambiguity, and an upgrade pass that re-examined Ambivalent predictions for directness. Crucially, only medium- and low-confidence predictions triggered re-examination, reducing API calls by approximately half while maintaining strong accuracy. This selective verification design is a pragmatic and effective engineering contribution.
CSE-UOI \cite{tzouvaras2026cseuoisemeval2026task6}, in contrast, relied on \textbf{combining the outputs of two different models} (Grok 4.1 and Gemini 3 Flash). Each model produced several candidate answers, which were then aggregated through a weighted voting scheme. Instead of re-querying the models, their system applied a second step that identifies difficult or ambiguous cases using simple signals (such as how long the model’s responses are or how consistent they are across attempts) and adjusts the final prediction accordingly. In this sense, while AsymVerify selectively re-checks uncertain cases by asking the model again, CSE-UOI handles uncertainty by combining multiple answers upfront and refining the decision without additional model calls.

Among LLM-based systems, \textbf{few-shot prompting combined with Chain-of-Thought (CoT)} reasoning was the near-universal strategy among top performers, and the specific prompt pipeline design was the primary differentiator between systems. Approaches included dynamically assembled prompts, iterative prompt repair loops, debate-style prompting, and CoT distillation. Zero-shot  prompting without further augmentation and standard task-instruction-only prompting were the weakest strategies, consistently absent from the top ranks.

\subsubsection{Joint vs. Separate Framing}
A key strategic decision among participants was whether to treat Subtask 1 \textbf{independently} or to couple it with Subtask 2. Among the top seven systems, the majority adopted a joint framing, four of which derived their Subtask 1 predictions by first predicting the Subtask 2 evasion label and then mapping it upward through the taxonomy. 

The inference is that the hierarchical taxonomy provides a natural intermediate supervision signal, and models that leveraged this structure obtained a systematic advantage. Conversely, teams that treated the tasks independently tended to cluster in the 0.72–0.80 range, with some exceptions which compensated through sophisticated multi-pass reasoning and hybrid architectures respectively.

\subsubsection{Fine-Tuned Encoders and Prompting Strategies}
Approximately 14 teams, clustering at the 0.75 range, relied primarily or exclusively on fine-tuned encoder-only models, chiefly RoBERTa and DeBERTa variants. These systems shared a common toolkit: weighted cross-entropy or focal loss to address class imbalance, minority class oversampling, stratified k-fold cross-validation, and ensembling across multiple fine-tuned models or random seeds. The best purely encoder-based system scoring at 0.81 was achieved only through \textbf{substantial architectural augmentation} combining NLI reframing, multi-task learning, Graph Neural Networks (GNNs), and Multiple Instance Learning,  an indication that standard encoder fine-tuning alone has largely reached its ceiling on this subtask.

\subsubsection{Class Imbalance}
The Ambivalent class being by far the most frequent in the data was a widely acknowledged challenge addressed through several complementary strategies: weighted loss functions, minority class oversampling, domain-adaptive pretraining on the interview corpus, and LLM-generated synthetic data augmentation reaching into the thousands of additional training examples. For LLM-based systems, imbalance awareness was sometimes encoded directly into prompts through explicit base-rate priors and skepticism instructions calibrated to the empirical class distribution, effectively substituting prompt design for loss reweighting.

\subsection{Subtask 2 - Evasion-level}
\subsubsection{Overview and Participation}
Subtask 2 required participants to classify each QA pair into one of nine fine-grained evasion categories organised under the CLARITY taxonomy. Nineteen teams submitted results for this subtask, which proved \textbf{substantially harder} than Subtask 1: the highest score was 0.68 Macro-F1, surpassing the provided baseline, compared to 0.89 in Subtask 1, and the median score was approximately 0.52. This score compression reflects the inherent difficulty of distinguishing nine semantically overlapping evasion categories from a relatively small dataset. The complete leaderboard is shown in Table \ref{tab:leaderboards}.
\begin{table*}[t]
\centering
\small
\begin{tabular}{clc|clc}
\hline
\multicolumn{3}{c|}{\textbf{Subtask 1: Clarity Level}} &
\multicolumn{3}{c}{\textbf{Subtask 2: Evasion Level}} \\
Rank & Team & Macro-F1 & Rank & Team & Macro-F1 \\
\hline
1 & TeleAI & 0.89 & 1 & TeleAI & 0.68 \\
2 & AsymVerify & 0.85 & 2 & moswisarut & 0.61\\
3 & CSE-UOI & 0.85 & 3 & CLaC @ CLARITY & 0.59\\
4 & Rasende Rakete & 0.83 & 4 & Rasende Rakete & 0.59  \\
5 & Evaluators & 0.83 & 5 & YNU-HPCC & 0.59\\
6 & YNU-HPCC & 0.83 & 6 & pressprexx & 0.58\\
7 & moswisarut & 0.82 & 7 & CSE-UOI & 0.58  \\

\rowcolor{lightblue}
 & Best Baseline (Llama 70 B) & 0.82 &  & Best Baseline (Llama 70B) & 0.57 \\

8 & tahamunawar & 0.81 & 8 & ttda704 & 0.56\\
9 & CLaC @ CLARITY & 0.80 & 9 & Evaluators & 0.54\\
10 & SpinDetector & 0.80 & 10 & gsdeyson & 0.52\\
11 & gabriel\_stefan & 0.80 & 11 & gabriel\_stefan & 0.51 \\
12 & AGAI & 0.79 & 12 & taleef & 0.50\\
13 & ttda704 & 0.79 & 13 & KCLarity & 0.50\\
14 & silkpeak & 0.76 & 14 & AI@UMS & 0.48\\
15 & CUET\_823 & 0.76 & 15 & JeisonJimenez & 0.47\\
16 & aset\_clarity & 0.76 & 16 & uir\_cis & 0.47\\
17 & goru18 & 0.76 & 17 & tahamunawar & 0.45 \\
18 & taleef & 0.76 & 18 & aliraza1245 & 0.45  \\
19 & Shuja & 0.76 & 19 & 3devs & 0.44 \\
20 & argha & 0.75 & 20 & arsalaaan & 0.43\\
21 & The Argonauts & 0.75 & 21 & danhle32 & 0.43\\
22 & KCLarity & 0.74 & 22 & The Argonauts & 0.42\\
23 & Limit Testing Special Forces & 0.74 & 23 & sammywal & 0.42\\
24 & SyntaxMind & 0.72 & 24 & aliraza4353 & 0.41\\
25 & gsdeyson & 0.72 & 25 & FER- Akrap-Bilic-Cuturilo-Racic-Simpraga & 0.40\\
26 & tbernal & 0.71 & 26 & viettien & 0.39\\
27 & rafsan & 0.68 & 27 & fridlowski & 0.39\\
28 & viettien & 0.68 & 28 & saifsaif & 0.34\\
29 & mikebeth & 0.68 & 29 & lakksh & 0.33\\
30 & JeisonJimenez & 0.68 & 30 & Shuja & 0.29 \\
31 & csecudsg & 0.67 & 31 & ziaulrehman43 & 0.28 \\
32 & jeffreyzhao & 0.64 & 32 & krishna11098 & 0.25 \\
33 & B\&B & 0.64 & 33 & yx-ym & 0.13 \\
34 & AI@UMS & 0.62 &  &  &  \\
35 & uir\_cis & 0.61 &  &  &  \\
36 & 3devs & 0.59 &  &  &  \\
37 & AI@UMS & 0.59 &  &  &  \\
38 & krishna11098 & 0.43 &  &  &  \\
39 & Happy frogs & 0.42 &  &  &  \\
40 & lakksh & 0.31 &  &  &  \\
41 & yx-ym & 0.28 &  &  &  \\
\hline
\end{tabular}
\caption{Leaderboard results for both subtasks ranked by Macro-F1 score. Full leaderboard available at
\href{https://konstantinosftw.github.io/CLARITY-SemEval-2026/\#leaderboard}{task website}.}
\label{tab:leaderboards}
\end{table*}

\subsubsection{Dominant Approaches: Three-Stage Pipeline \& Hybrid Encoder-LLM}
The winning system for Subtask 2 was TeleAI (0.68), the same team that achieved first position Subtask 1. Their approach is also the most architecturally sophisticated in the competition. Rather than directly classifying all nine categories in a single pass, they decomposed the problem into \textbf{three sequential stages}. Stage 1 classified each sample into one of four macro-categories: Clarification, Claims Ignorance, Declining to Answer, or OTHER, including an attached verbalized confidence score. Stage 2 was triggered for the three Non-Reply categories when their confidence was medium or low, and used dynamically assembled prompts (shared template plus label definitions, label-specific confusion guidelines, and few-shot boundary examples selected based on the Stage 1 label) to re-examine and finalize the prediction across all nine classes. Stage 3 handled the OTHER samples from Stage 1, applying the same dynamic construction framework but switching to six-class definitions covering the valid-answer evasion types.

The key design insight is that the pipeline routes examples based on confidence and coarse category, ensuring that the most confusable cases receive the most contextualised prompting. This confidence-conditional routing significantly reduces the probability of systematic confusion between semantically adjacent categories (e.g., Dodging vs. Deflection, or Claims Ignorance vs. Declining to Answer), which is the primary failure mode for simpler one-pass classifiers. The performance gap between the first (0.68) and the second-placed team (0.61) is the largest single gap in the Subtask 2 leaderboard, underscoring the efficacy of this architecture.

The second-ranked system employed a conceptually elegant \textbf{hybrid approach} that merits particular attention. The system first fine-tuned RoBERTa-large to produce a ranked list of the top five most likely evasion categories per QA pair. This ranked candidate set was then passed as a constrained label space to Kimi-K2, which performed few-shot inference using exactly the examples provided in the original CLARITY task paper, conditioned on the five candidates rather than the full nine-class set. The clarity label was then inferred hierarchically from the predicted evasion type.

This architecture effectively uses the fine-tuned encoder as a \textbf{recall-oriented first-pass filter} to reduce the output space for the LLM, while the LLM provides \textbf{precision-oriented discrimination} among the shortlisted candidates. The result is a system that is computationally more efficient than full nine-class LLM prompting while avoiding the accuracy ceiling of encoder-only classification. Achieving rank 2 overall while being the only system in the top three to use a local GPU (Google Colab L4, with a training time under 1.5 hours) makes this one of the most resource-efficient high-performing systems in the competition.

\subsubsection{Hierarchical Decomposition \& Encoder Limitations}
A recurring and statistically meaningful pattern among top-performing Subtask 2 systems was the explicit \textbf{exploitation of our taxonomy's hierarchical structure}. Rather than treating the problem as flat nine-class classification, the most successful designs decomposed it into sequential or partitioned subproblems aligned with the taxonomy's branching logic. Instantiations of this principle varied in granularity: some systems first predicted the coarse Subtask 1 clarity label and then restricted inference to the corresponding taxonomy branch, effectively reducing the output space to two or three candidates per branch; others routed examples through confidence-conditional stages, applying increasingly targeted prompts only to cases where prior stages expressed uncertainty; and others assigned the easiest class (Clear Replies, which deterministically map to the Explicit evasion type) without further inference, concentrating model capacity entirely on the genuinely ambiguous cases. Across all variants, the consistent finding was that systems implementing any form of hierarchical decomposition outperformed those attempting direct nine-class inference, strongly suggesting that flat classification is structurally ill-suited to this taxonomy.

Fine-tuned encoder-only models fared considerably worse in Subtask 2 than in Subtask 1, with the best purely encoder-based result reaching only 0.50, a gap of 0.18 Macro-F1 pp below the top system. Even heavily engineered encoder pipelines, including large multi-seed ensembles of DeBERTa-v3-large, teacher-to-student knowledge distillation via QLoRA, architectures combining NLI reframing, multi-task learning, GNNs, and Multiple Instance Learning, all fell below 0.51 and were consistently outperformed by simpler LLM-prompted approaches. The semantic proximity of several evasion categories, particularly among Dodging, Deflection, General, and Implicit, appears to require the broader world knowledge and generative reasoning capacity of large language models to resolve, and no amount of ensemble complexity or architectural augmentation within the encoder paradigm was sufficient to bridge this gap.

One notable exception to the binary of pure prompting versus pure fine-tuning was the use of \textbf{supervised fine-tuning augmented with CoT distillation}: training open-weight models of moderate size (in the 4B–14B range) on synthetic reasoning traces generated by a larger model, combined with structured prompts that walked the model through the taxonomy levels step by step. This approach achieved results competitive with multi-agent systems built on proprietary frontier models, demonstrating that a 14B parameter open-weight model with distilled reasoning supervision can match the discriminative capacity of much larger closed-weight systems on this task. Given its reproducibility, low inference cost, and independence from API access, CoT distillation into mid-sized open-weight models represents a particularly promising direction for future participation.

\subsubsection{Class Imbalance}
Class imbalance is more acute in Subtask 2 than Subtask 1, with nine categories to distinguish from approximately 3,400 training examples. Standard mitigation strategies — focal loss, inverse-frequency class weighting, and minority class oversampling — were widely applied but proved insufficient on their own to close the performance gap. More aggressive data-centric approaches included generating thousands of synthetic training examples using frontier LLMs, fine-tuning models across a wide parameter range (1B to 70B) on synthetic CoT reasoning data produced by a very large teacher model, and applying layer-frozen encoder fine-tuning with extensive cross-validation. While these approaches did not produce top-ranked submissions, they collectively reflect significant interest in addressing the small-dataset bottleneck and suggest that larger or more carefully curated training data could meaningfully shift the performance ceiling for this subtask.

\subsubsection{Score Compression and Task Difficulty}
The upper bound of 0.68 Macro-F1 in Subtask 2, compared to 0.89 in Subtask 1, reflects several compounding sources of difficulty. First, the nine-class output space creates much more opportunity for \textbf{inter-class confusion} than a three-class problem. Second, several of the evasion categories are \textbf{semantically proximal} (e.g., Dodging and Deflection, Implicit and General), requiring precise contextual understanding to distinguish. Third, the training set size is \textbf{modest} relative to the output space complexity. Fourth, annotation agreement for fine-grained categories is inherently lower than for coarse clarity labels, which may introduce \textbf{label noise} that depresses achievable performance.

The clustering of scores between 0.47 and 0.61 for ranks 3–17, spanning a range of only 0.14 Macro-F1 over 15 teams, further underscores how competitive and tightly constrained the performance frontier is for this subtask. Even the most sophisticated systems leave substantial room for improvement, and the task remains an open challenge particularly for distinguishing the six valid-answer evasion subcategories.

\section{Discussion and Conclusion}
The CLARITY shared task attracted 30 participating teams and produced a rich and methodologically diverse set of systems for detecting clarity and evasion in political discourse. Across both subtasks, the results paint a consistent and instructive picture of the current state of the art for discourse-level classification tasks of this kind.

For Subtask 1, the most important finding is the substantial and systematic advantage of LLM-based approaches over fine-tuned encoder models. All top-ranked systems relied on LLMs with sophisticated prompting pipelines, and the winning strategy bypassed direct clarity classification entirely by first solving the harder fine-grained evasion task and mapping predictions back through the taxonomy hierarchy, a result that indicates something fundamental about the structure of the problem: coarse clarity labels are more reliably derived as a consequence of fine-grained evasion reasoning than they are learned directly from supervised signal alone. For Subtask 2, the nine-class evasion classification task proved considerably harder, with a performance ceiling of 0.68 Macro-F1 and a large cluster of systems between 0.45 and 0.61, reflecting the genuine semantic difficulty of distinguishing proximal evasion categories such as \textit{Dodging}, \textit{Deflection}, \textit{General}, and \textit{Implicit}. Hierarchical decomposition of the output space in any of its several instantiations was the single most consistent differentiator between strong and weak systems across both subtasks, confirming that the taxonomic structure of the task is not merely an organisational convenience but an active source of inductive bias that systems should exploit.

Several secondary findings merit emphasis for the wider research community. First, encoder-only models appear to have reached a practical ceiling on this task, with performance saturating around 0.76–0.81  Macro-F1 for Subtask 1 and 0.50 for Subtask 2 regardless of scale, ensembling, or architectural augmentation. Second, CoT distillation into mid-sized open-weight models proved to be a compute-efficient and reproducible alternative to proprietary API-based inference, achieving results competitive with frontier model systems at a fraction of the cost. Third, prompt design quality, including confidence-conditional routing, dynamic few-shot anchor selection, and iterative refinement, was a stronger determinant of LLM system performance than model size or model choice alone. Finally, class imbalance, while universally acknowledged, was most effectively handled not through loss reweighting alone but through structural interventions: injecting empirical priors directly into prompts, generating synthetic training examples, or routing easy-to-classify instances deterministically to avoid wasting model capacity on them.

Looking ahead, the challenging CLARITY task opens several future directions. The strong performance of hierarchical inference pipelines  suggests that future shared tasks in this space may benefit from explicit multi-stage evaluation protocols that reward intermediate predictions at each level of the taxonomy. Furthermore, the persistent gap between human-level understanding of political evasion and current system performance, particularly for fine-grained evasion classification, suggests that richer contextual representations of discourse, incorporating speaker history, interview dynamics, and rhetorical framing, could yield meaningful improvements. Multimodal extensions incorporating prosodic and gestural cues from video interview recordings represent a further avenue, as evasion in spoken political discourse is often signalled through paralinguistic features invisible to text-only systems.  More broadly, the CLARITY dataset and the systems developed for this task provide a foundation for studying political communication computationally across languages, political systems, and interview contexts, a direction of growing relevance as automated analysis of political discourse becomes increasingly consequential.

\section*{Limitations}
The CLARITY shared task constitutes, to our knowledge, one of the first community-wide efforts to operationalize political question evasion as a computational task. Nevertheless, some limitations should be acknowledged.

First, the dataset is restricted to \textit{English-language U.S. presidential interviews} collected from the official White House archive. Although this design choice ensures high-quality transcripts, a relatively controlled discourse setting and availability of annotators, it limits the generalizability of the findings to other languages and political systems. Evasive behavior is shaped by institutional norms, media formats, political cultures, and language-specific pragmatic conventions; therefore, systems developed for CLARITY may not transfer directly to other countries, political systems, or communicative arenas such as parliamentary debates, press briefings or campaign rallies.

Moreover, the annotation process itself necessarily involves \textit{subjective judgment}. The distinction between directness, ambiguity, and specific evasion strategies is inherently interpretive, especially in politically strategic discourse where multiple readings may plausibly coexist. While we attempted to mitigate this through annotator training, expert supervision, overlapping annotation, and agreement analysis, some amount of disagreement is not only unavoidable but conceptually intrinsic to the phenomenon under study. This limitation is particularly salient for the fine-grained taxonomy, where several label pairs exhibit lower agreement.

Finally, the benchmark focuses exclusively on \textit{textual} information. In real political interviews, audiences also rely on prosody, hesitation, timing, facial expressions, gestures, and interactional dynamics to infer whether a politician is evading a question. Because such multimodal signals are absent from the present task, the benchmark captures only the textual component of evasiveness and may underestimate complementary or contradictory cues available in natural settings.

\section*{Ethical Considerations}
CLARITY focuses on publicly available political interviews involving public figures acting in institutional roles. The purpose of the shared task is to support research on political communication, accountability, and language understanding, not to produce definitive judgments about a speaker's honesty, intent, or moral character. In this sense, the benchmark should be understood as a tool for analyzing \textit{response clarity} and \textit{evasion patterns} as discourse phenomena, rather than as a mechanism for certifying whether a politician is being truthful or deceptive.

At the same time, the task necessarily relies on human interpretation. Deciding whether a response is clear, ambiguous, or evasive is not a purely objective process, especially in political discourse where multiple plausible readings may coexist. Although we mitigated subjectivity through annotator training, expert supervision, and agreement analysis, the resulting labels should still be treated as structured human judgments under a specific taxonomy, not as indisputable ground truth. This is particularly important when interpreting borderline cases, where disagreement may reflect genuine ambiguity rather than annotation error.

A further ethical concern is the potential misuse of systems developed for this benchmark. Automatic predictions of evasion may be overinterpreted in public debate, journalism, or downstream monitoring tools, especially if they are presented without context or uncertainty. We therefore caution against using model outputs as stand-alone evidence about political actors. Any practical deployment should preserve human oversight, account for context, and avoid collapsing nuanced discourse behavior into simplistic or partisan conclusions.

\section*{Acknowledgments}
We sincerely thank our annotators for the time and effort they invested in providing high-quality annotations to the CLARITY dataset.


\bibliography{custom}

\appendix

\section{Human annotation}
\subsection{Fleiss $\kappa$}
\label{sec:fleiss}
By considering $N$=317 items, $n$=3 annotators per item, $k$ categories (3 for the clarity level, 9 for the evasion level) and $n_{ij}$ the number of annotators that assigned item $i$ to category $j$, the per-item agreement is calculated as:
\begin{equation}
 P_i = \frac{1}{n(n-1)} \sum_{j=1}^{k} n_{ij}(n_{ij} - 1),
\quad i = 1, \dots, N   
\end{equation}
The mean observed agreement is:
\begin{equation}
\bar{P} = \frac{1}{N} \sum_{i=1}^{N} P_i.
\end{equation}
The overall proportion of assignments to class $j$ is:
\begin{equation}
p_j = \frac{1}{Nn} \sum_{i=1}^{N} n_{ij}.
\end{equation}
The expected agreement by chance is:
\begin{equation}
\bar{P_e} = \sum_{j=1}^{k} p_j^2.
\end{equation}
Finally, Fleiss’ kappa is then computed as:
\begin{equation}
\kappa = \frac{\bar{P} - \bar{P_e}}{1 - \bar{P_e}}.
\end{equation}

This configuration of $\kappa$ is employed for evaluating annotators' responses.

\end{document}